# Computer Vision Systems in Road Vehicles: A Review

Kristian Kovačić, Edouard Ivanjko and Hrvoje Gold
Department of Intelligent Transportation Systems
Faculty of Transport and Traffic Sciences
University of Zagreb
Email: kristian.kovacic@zg.t-com.hr, edouard.ivanjko@fpz.hr, hrvoje.gold@fpz.hr

*Abstract*—The number of road vehicles significantly increased in recent decades. This trend accompanied a build-up of road infrastructure and development of various control systems to increase road traffic safety, road capacity and travel comfort. In traffic safety significant development has been made and today's systems more and more include cameras and computer vision methods. Cameras are used as part of the road infrastructure or in vehicles. In this paper a review on computer vision systems in vehicles from the stand point of traffic engineering is given. Safety problems of road vehicles are presented, current state of the art in-vehicle vision systems is described and open problems with future research directions are discussed.

## I. Introduction

Recent decades can be characterized by a significant increase of the number of road vehicles accompanied by a build-up of road infrastructure. Simultaneously various traffic control systems have been developed in order to increase road traffic safety, road capacity and travel comfort. However, even as technology significantly advanced, traffic accidents still take a large number of human fatalities and injuries.

Statistics show that there is about 1.2 million fatalities and 50 million injuries related to road traffic accidents per year worldwide [1]. Also there were 3 961 pedestrian fatalities in 2005 in the EU. Most accidents happened because drivers were not aware of pedestrians. Furthermore, there has been approximately 500 casualties a year in EU because of driver's lack of visibility in the blind zone. Data published in [1] estimated that there has been around 244 000 police reported crashes and 224 fatalities related to lane switching in 1991. Statistics published by the Croatian Bureau of Statistics [2] reveal that in 2011 there were 418 killed and 18 065 injured persons who were participating in the traffic of Croatia.

To reduce the amount of annual traffic accidents, a large number of different systems has been researched and developed. Such systems are part of road infrastructure or road vehicles (horizontal and vertical signalization, variable message signs, driver support systems, etc.). Vehicle manufactures have implemented systems such as lane detection and lane departure systems, parking assistants, collision avoidance, adaptive cruise control, etc. Main task of these systems is to make the driver aware when leaving the current lane. These systems are partially based on computer vision algorithms that can detect and track pedestrians or surrounding vehicles [1].

Development of computing power and cheap video cameras enabled today's traffic safety systems to more and more include cameras and computer vision methods. Cameras are used as part of road infrastructure or in vehicles. They enable monitoring of traffic infrastructure, detection of incident situations, tracking of surrounding vehicles, etc.

The goal of this paper is to make a review of existing computer vision systems from the problem stand point of traffic engineering unlike the most reviews done from the problem stand point of computer science. Emphasis is on computer vision methods that can be used in systems build in vehicles to assist the driver and increase traffic safety.

This paper is organized as follows. Second section gives a review of road vehicles safety problems followed by the third section that describes vision systems requirements for the use in vehicles. Fourth section gives a review of mostly used image processing methods followed by the fifth section that describes existing systems. Paper ends with a discussion about open problems and conclusion.

## II. Safety problems of road vehicles

According to the survey study results published in [3], annually about 100 000 crashes in the USA result directly from driver fatigue. It is the main cause of about 30% of all severe traffic accidents. In the case of French statistics, a lack of attention due to driver fatigue or sleepiness was a factor in one in three accidents, while alcohol, drugs or distraction was a factor in one in five accidents in 2003 [4].

Based on the accidents records from [3] most common traffic accidents causes are:

- Frontal crashes, where vehicles driving in opposite directions have collided;
- Lane departure collisions, where a lane changing vehicle collided with a vehicle from an adjacent lane;
- Crashes with surrounding vehicles while parking, passing through a intersection or a narrow alley, etc.;
- Failures to see or recognize the road signalization and consequently cause a traffic accident due to inappropriate driving.

### A. Frontal crashes

According to the data published in [3], 30 452 of total 2 041 943 vehicles crashes recorded in the USA in the period from 2005 to 2008 occurred because of too close following of a vehicle in a convoy. About 5.5% of accidents happened with vehicles moving in opposite direction. Although there are various types of systems that can help to reduce the number of accidents caused by a frontal vehicle crash, computer vision gives ability to prevent a possible or immediate vehicle crash from happening. System can detect a dangerous situation using one or more frontal cameras and afterward produce appropriate response, haptic or audio.

When a possible frontal crash situation occurs, the time available to prevent the crash is usually very short. Efficiency of crash avoidance systems depends on the time in which possible crash has been detected prior to its occurrence. If short amount of time is taken for possible frontal crash detection, then longer amount of time is available for driver warning or evasive actions to be performed. Such actions need to be taken in order to prevent or alleviate the crash (preparation of airbags, autonomous braking, etc.).

*B. Lane departure*

Statistics published in [3] denote that 10.8% accidents happened because vehicles overran the lane line and 22.2% because vehicles were over the edge of the road. From all vehicles that participated in accidents, 3.1% were intentionally changing lanes or overtaking another vehicle.

Number of accidents can be significantly reduced by using a computer vision system for lane departure warning. If the driver is distracted, such system can simply just accentuate the dangerous action (lane changing) and focus the driver's attention. When such system is built into a vehicle, drivers also tend to be more careful so that they do not cross the lane markers because they receive some kind of warning otherwise (vibrating driver seat, audio signal or graphical notice).

*C. Surrounding vehicles*

Surrounding vehicles represent a threat if the driver is not aware of them. In case of driver fatigue, capability of tracking surrounding vehicles is significantly reduced. This can cause an accident in events such as lane changing, driving in a convoy, parking and similar situations.

Depending on complexity of a system that tracks surrounding vehicles, appropriate warnings and actions can be performed based on current vehicle environment state or state that is estimated (predicted) to happen in the near future. System that tracks surrounding vehicles can be used in lane departure warning or parking assistant system also. Main problem of this system are false detections of surrounding vehicles. Vehicles can be in many different shapes and colors. This can cause false vehicle estimations and wrong driver actions endangering other drivers and road users [5].

*D. Recognition of road signalization*

Road signalization (horizontal and vertical) represents the key element in traffic safety. Driver should always detect and recognize both road signalization groups correctly, especially traffic signs. In case when the driver is unfocused or tired, horizontal and vertical signalization is mostly ignored. According to the survey [3], 3.8% of total vehicle crashes happened when drivers did not adjust their driving to road signalization recommendation. This accidents number could be reduced if a convenient road signalization detection and driver assistant system had been used. Such system could warn the driver when he is performing a dangerous or illegal maneuver. According to data in [3], there were 109 932 accidents caused by speeding and ignoring speed limits.

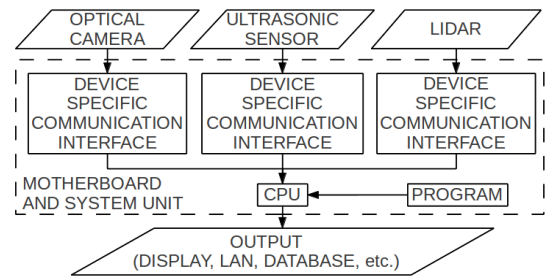

Fig. 1. General in-vehicle driver support system architecture.

In some cases drivers are overloaded with information and have a hard time prioritizing information importance. For example, when driving through an area for the first time, drivers are usually more concentrated on route information traffic signs. Systems for road sign detection and recognition, can improve driving safety and reduce above mentioned accidents percentage by drawing the driver attention to warning signs.

### III. VISION SYSTEMS REQUIREMENTS

To define an appropriate vision system architecture, some application area related requirements need to be considered. Because of the specific field of use, the computer vision system has to work in real time. Obtained data should be available in time after a possible critical situation has been detected. Time window for computation depends on the vehicle speed and shortens with the vehicle speed increase. This can represent a significant problem when small and cheap embedded computers are used. To overcome this problem algorithms that support parallel computing, multi-core processor platforms or computer vision dedicated hardware platforms are used. Alternatively special intelligent video cameras with preinstalled local image processing software can be used. Such cameras can perform extraction of basic image features, object recognition and tracking.

Second important requirement is the capability to adapt to rapid changes of environment monitored using cameras. Such changes can be caused by fog, rain, snow, and illumination changes related to night and day or entering and exiting tunnels. In cases where whole driver support or vehicle control systems use more then one sensor type (eg. optical cameras and ultrasonic sensors, lidars), system adaptiveness to environment changes is better then in cases with one sensor type. Furthermore, whole system needs to be resistant to various physical influences (e.g. vibrations, acceleration/deceleration) [5]. In Fig. 1 general in-vehicle vision system architecture is given.

To detect and track vehicles or other kind of objects (pedestrians, lane markings, traffic signs, etc.) often different sensors are used. Sensors can be divided into two categories: active and passive. Most active sensors are time of flight sensors, i.e. they measure the distance between a specific object and the sensor. Millimeter-wave radars and laser sensors are mostly used as active sensors in automotive industry. Active sensors are so mainly used for surrounding vehicles detection and tracking.

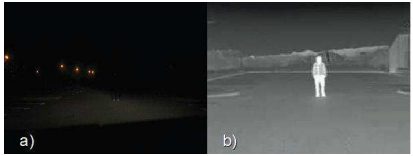

Fig. 2. Same image in visible spectrum (a) and in far-infrared spectrum (b).

Various cameras are mostly used as passive sensors. Basic camera characteristics include their working spectrum, resolution, field of view, etc. In traffic application camera spectrum is crucial since night-time cameras use far, mid and near infrared spectrum, while daylight cameras use visible spectrum. Far infrared spectrum camera is suitable for detection of pedestrians at night as shown in Fig. 2.

Computer vision systems in vehicles can be based on stereo vision also. Stereo vision is a technique that enables extraction of 3D data from a pair of 2D flat images. In case of a 2D image pair, usually whole overlapping image or only a specific region of it is translated from a 2D to 3D coordinate system.

## IV. Image processing approaches

Image processing consists of basic operations that provides features for high level information extraction. Simplest image processing transformations are done by point operators. Such operators can perform color transformation, compositing and matting (process of separating the image into two layers, foreground and background), histogram equalization, tonal adjustment and other operations. Point operators do not use surrounding points in calculations. Contrariwise, neighborhood operators (linear filtering, non-linear filtering, etc.) use information from surrounding pixels also. For image enhancement, linear and non-linear filters (moving-average, Bartlett filter, median filter, morphology, etc.) are used. Non-linear filters can be used for shape recognition methods also. For example, as part of vehicle recognition methods [6].

Affine transformation is a type of transformation that preserves symmetries (parallelism) and distance ratios when applied on certain points or parts of an image. As a consequence of this, specific geometric shape on an image can be rotated, scaled, translated, etc. This transformation is usually used for basic operations with a whole specific image region.

Distance estimation between vehicles is important in traffic applications. To calculate the distance between the camera and a point in space represented by a specific pixel in the image, perspective (projection) transformation is performed. It is computed by applying a perspective matrix transformation on every pixel in the image. Perspective transformation assigns a third component (z coordinate) to every pixel depending on the x-y components in the image. In Fig. 3, result of perspective image transformation is given. Pixel components in the original image are x and y coordinates. After perspective transformation a new pixel component (z coordinate) is calculated and swapped with the y coordinate. So, pixels near the image bottom of the image represent points in 3D space that are closer to the camera and pixels near the image top are further away.

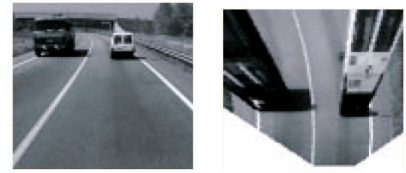

Fig. 3. Original image (left) and same image with perspective projection transformation applied (right) [7].

Hough transformation is used to extract parameters of image features (lines). Main advantage of this method is its relatively high tolerance to noise and ability to ignore small gaps on shapes. After required processing (edge detection) is done on an image, Hough transformation translates every pixel of a feature in the image from the Cartesian to the polar coordinate system. Original Hough transform method is used for extracting only line parameters although various modified versions of Hough transform have been developed to extract corner, circle and ellipse parameters.

For object detection Paul Viola and Michael Jones proposed an approach called 'Haar-like features' that has some resemblance to Haar-wavelet functions [8]. Haar-like-features methods are used where fast object detection is required [9], [10]. This approach is based on dividing a detection window (window of interest) into smaller regions. Region size depends on the size of an object that needs to be detected. For each region a sum is computed based on pixel intensity within the processed region. Summed area table (integral image) is used to perform a fast summarization. Possible object candidates can be found comparing differences between sums of adjacent regions. For a more accurate object detection, many regions need to be tested. Applications such as vehicle detection require use of more complex methods that have computational demands similar to the Haar-like-features method.

One of the algorithms used for simplification of extracted shapes is the Ramer-Douglas-Peucker algorithm [11]. It reduces the number of points in a curve that is approximated by a series of points. Complex curves with many points can be too computationally (time) demanding for processing. Simplifying the curve with this algorithm allows faster and simpler processing.

## V. Existing road vehicle computer vision systems

Since the first intelligent vehicle was introduced in the mid 1970s [12], computer vision systems also started to develop as one of its main sensor systems. Currently there are many projects that are researching and developing computer vision systems in road vehicles. Project goals are related to development of driver support systems, driver state recognition and road infrastructure recognition modules for autonomous vehicles.

Computer vision systems in vehicles are part of autonomous vehicle parking systems, adaptive cruise control, lane departure warning, driver fatigue detection, obstacle and traffic sign detection, and others. Currently only high class vehicles are equipped with several such systems but they are being more and more included in middle class vehicles.

*A. Lane detection and departure warning*

Lane detection system represents not only a useful driver aid, but an important component in autonomous vehicles also. To resolve lane detection and departure warning problems, key element is to detect road lane boundaries (horizontal road signalization and width of the whole road). In work described in [13], lane detection is done in two parts. First part includes image enhancement and edge detection. Second part deals with lane features extraction and road shape estimation from the processed image. In Fig. 4, more detailed flow diagram of this algorithm is shown.

Authors of [14] developed an algorithm that handles extraction of lane features based on estimation methods. Besides determining shape of main (current) lane, algorithm has the ability to estimate position and shape of other surrounding lanes. Applied lane detection principle is based on two methods. On a straight lane, Hough and simplified perspective transformations mentioned in previous section are used first. These transformations enable detection of the center lane. To detect surrounding lanes, further processing needs to be done. Required processing includes affine transformations and other calculations that retrieve relative vehicle position. If the lane is a curve with a large radius, this method can be used also. In second method for curved lane with small radius, edge features are extracted from the original image first. After extracting these features, complete perspective transformation is performed based on camera's installation parameters. By recognizing points on the center lane, curve estimation is performed. Positions of other adjacent lanes are then calculated based on known current lane curve parameters. Problems with this approach are: (i) incapability to detect lane markers due the poor visibility or object interference (infront vehicles or other objects block the camera's field of view); and (ii) error in estimating current vehicle position due to the curvature of the road [14].

*B. Driver fatigue detection*

The process of falling asleep at the steering wheel can be characterized by a gradual decline in alertness from a normal state due to monotonous driving conditions or other environmental factors. This diminished alertness leads to a state of fuzzy consciousness followed by the onset of sleep [15]. If drivers in fuzzy consciousness would get a warning one second before possible accident situation, about 90% of accidents could be prevented [9]. Various techniques have been developed to help to reduce the number of accident caused by driver fatigue. These driver fatigue detection techniques can be divided into three groups: (i) methods that analyze driver's current state related to eyelid and head movement, gaze and other facial expression; (ii) methods based on driver performance, with a focus on the vehicle's behavior including position and headway; and (iii) methods based on combination of the driver's current state and driver performance [15].

In work described in [9], two CCD cameras are used. First camera is fixed and it is used to track the driver's head position. Camera output is an image in low-resolution like 320x240, which allows the use of fast image processing algorithms. As the driver's head position is known, processing of second camera image can focus only on that region. Image of second camera is high-resolution one (typically 720x576) which raises level of detail of the driver's head. This image is used in further analysis for determining driver's performance from mouth position, eye lid opening frequency and diameter of opened eye lid [9]. It is visible from experimental results that at least 30 frames (high-resolution images) can be processed per second. Such frame rate represents good basis for further research and development of driver fatigue detection systems.

In review [15], an approach to driver fatigue detection that first detects regions of eyes and afterwards extracts information about driver fatigue from this region is briefly described. To achieve this, driver's face needs to be found first. After finding the face, eyes locations are estimated by finding the darkest pixels on face. When eye locations are known, software tries to extract further information for driver fatigue detection. In order to test the algorithm different test drives were made. If the driver's head direction was straight into the camera none false detection were observed. After head is turned for 30 or more degrees, system starts to produce false detections. Beside described techniques, processing additional features like blinking rate of driver's eyes or position of eye's pupil are used to improve driver fatigue detection [9].

*C. Vehicle detection*

There are several sensors (active and passive) that can be used for vehicle detection. Lot of research was done regarding vehicle detection using optical cameras and this still presents a challenge. Main problem of video image processing based vehicle detection is significant variability in vehicle and environment appearance (vehicle size and shape, color, sunlight, snow, rain, dust, fog, etc.). Other problems arise from the need for system robustness and requirements for fast processing [7].

Most algorithms for vehicle detection in computer vision systems are based on two steps: (i) finding all candidates in an image that could be vehicles; and (ii) performing tests that can verify the presence of a vehicle. Finding all vehicle candidates in an image is mostly done by three types of methods: (i) knowledge-based methods; (ii) stereo-vision methods; and (iii) motion-based methods.

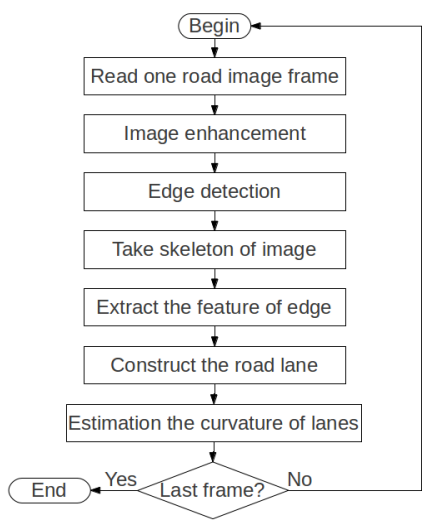

Fig. 4. Flowchart of image processing for lane detection [13].

Knowledge-based methods work on the presumption of knowing specific parameters of the object (vehicle) being detected (vehicle position, shape, color, texture, shadow, lighting, etc.). Symmetry can be used as one of the most important vehicle specifications. Every vehicle is almost completely symmetrical viewed from front and rear side and this is used in vehicle detection algorithms. Besides symmetry, color of the road or lane markers can be used in vehicle detection also. Using color, moving objects (vehicles) can be segmented from the background (road). Other knowledge-based methods use vehicle shadow, corners (vehicle rectangular shape), vertical/horizontal edges, texture, vehicle lights, etc. [7].

Stereo-vision methods are based on disparity maps or inverse perspective mapping. In disparity maps, position of pixels on both cameras that represents the same object are compared. Third component of pixel vector (z coordinate) can be calculated by comparing changes in pixels position. Inverse perspective mapping method is based on transforming the image and removing the perspective effect from it. In stereo-vision systems mostly two cameras are used although systems with more than two cameras are more accurate, but more computationally expensive also. In [16], a computer vision system with three cameras is described. It can detect objects that are 9 $[cm]$ in size and up to 110 $[m]$ in distance. Such a system has a high level of accuracy when compared to a system with two cameras.

Motion-based methods use optical flow for image processing. Optical flow in computer vision represents the pattern of visible movement of objects, surfaces and edges in the camera field of view. Optical flow calculated from an image is shown in Fig. 5, where arrow's length and direction represents average speed and direction (optical flow) of subimages. Algorithms for this method divide the image into smaller parts. Smaller image parts are analyzed with previously saved images in order to calculated its relative direction and speed. With this method static and dynamic image parts can be separated and additionally processed [17].

Once optical flow of sub-images is calculated, moving objects (vehicles) candidates can be recognized. A verification method decides which candidates will be referenced as real objects. Verification is done by calibrating a speed threshold on the most efficient value. With this filtering, false detections are reduced. For further verification, methods that are also used in knowledge based vehicle detection methods (appearance and template based) are used. These kind of methods recognize vehicle features like edges, texture, shadow, etc.

*D. Real time traffic sign detection*

Traffic sign detection is performed in two phases. First phase is usually done by color segmentation. Since traffic signs have edges of a specific color, color segmentation represents a good base for easier extraction of traffic signs from image background. In [10], color-based segmentation is done in two steps: color quantization followed by region of interest analysis. For color model, RGB color space is usually used, although YIQ, YUV, L*a*b and CIE can be used also. After color segmentation, shape-based segmentation is performed for final detection of circles, ellipses and triangles. Second phase is traffic sign classification where various methods like template matching, linear discriminant analysis, support vector machine (SVM), artificial neural networks, and other machine learning methods can be used [10]. For traffic sign detection AdaBoost [18] and SURF [19] algorithms are used also.

Traffic sign detection approach described in [11] performs recognition in two main steps. First step is related to a traffic sign detection algorithm whose goal is to localize a traffic sign in the image with minimum noise. For detection, image preprocessing with color-based method threshold is used first. This method is fast and often suitable for use in real-time systems. After preprocessing, traffic sign approximation is performed with Ramer-Douglas-Peucker algorithm. On localized traffic sign image, preprocessing is performed and image is converted to binary image with resolution of 30x30 pixels. Second step represents the traffic sign recognition based on SVM, often C-SVM with linear kernel.

The developed traffic sign detection and recognition system described in [20] is based on a video sequence analysis rather than a single image analysis. Authors use a cascade of boosted classifiers of Haar-like-features (usually pixel intensity value) method to detect traffic signs. OpenCV framework is used to alleviate the implementation. Main disadvantages of traffic sign detection based on a Haar-like-features method boosted classifier cascade are: (i) poor precision compared to the required precision; and (ii) poor traffic sign localization accuracy where its true location (position in the image) is in most cases deviated. Framework for detecting, tracking and recognizing traffic signs proposed in [20] is shown in Fig. 6.

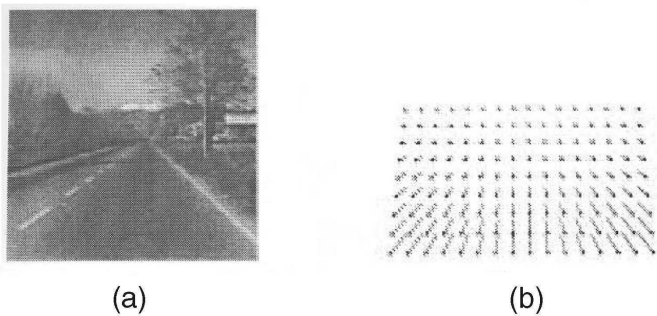

Fig. 5. Original image (a) and computed optical flow (b) [7].

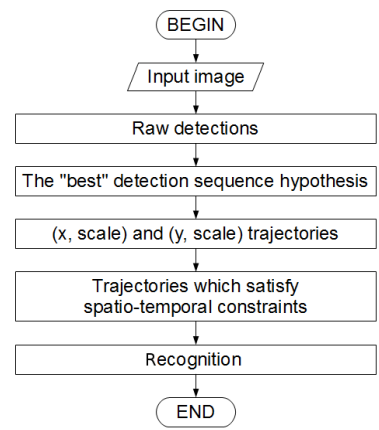

Fig. 6. Flowchart of sign detection and recognition algorithm [20].

Since traffic sign detection is performed on a sequence of images obtained from a moving vehicle, a region trajectory where most probably the signs are located can be constructed. Detection system learns on each detection. If a significant number of learning samples (traffic sign images) is used, system can achieve precision of up to $54\%$ with recall of $96\%$.

## VI. Open problems

Main constrain of computer vision systems in vehicles are their limitations due to hardware specifications. Processing an image through a large number of consecutive methods for data recognition and extraction can be very CPU demanding. Trade off between quality and cost of every step in image processing should be done in order to make a system maximally optimized. In image processing, two most widely used approaches are basic image features extraction using Hough transformation and learning classificators for whole object recognition. Implementation of classificators like the one based on Haar-like-features methods allows algorithms to use less system resources. Drawback of this method is its low ratio of accuracy. Contrary to this, Hough method is much more precise, but uses more system resources. Methods that perform complex calculations on all image pixels should be avoided and used only where they are necessary.

One of the factors that affects vehicle detection accuracy, tracking and recognition of objects in an outdoor environment are environmental conditions. Weather conditions such as snow, rain, fog and others can significantly reduce computer vision system accuracy. Poor quality of horizontal road signalization and omitted field of view due to various obstacles (often problem with traffic signs) presents an additional problem to vision based systems also. Significant robustness is required to overcome this problem. To reduce the influence that these factors can have on the system, new cameras and appropriate image processing algorithms still need to be developed.

## VII. Conclusion

Computer vision software used for vehicle detection and obtained recognition consist of many subsystems. Incoming images from one or several cameras need to be preprocessed in the desired format. Obtained image is then further processed for high level object detection and recognition.

Detecting an object in a scene can be done by multiple methods (eg. Hough and Haar-like-features methods). Use of this methods alone is not enough to make the system reliable for object detection and recognition. In cases of bad environment conditions, image can be full of various noise (snow, fog and rain on the image). Besides environment conditions, high vehicle velocity can also affects the image (blur effect).

Current systems that are used in more expensive vehicles are part of driver support systems based on simplified image processing. Examples are lane detection and traffic signs for speed limitations. Still problems related to robustness to different outdoor environment conditions and real time processing of high level features recognition exist.

## Acknowledgment

Authors wish to thank Nikola Bakarić for his valuable comments during writing this paper. This research has been supported by the project VISTA financed from the Innovation Investment Fund Grant Scheme and EU COST action TU1102.